\documentclass{article}
\usepackage{spconf,amsmath,graphicx}
\usepackage{subcaption}
\usepackage{booktabs}
\usepackage{times}
\usepackage{epsfig}
\usepackage{graphicx}
\usepackage{amsmath}
\usepackage{amssymb}
\usepackage{subcaption}
\usepackage{enumitem}
\usepackage{tablefootnote}
\usepackage{multirow}
\usepackage{hyperref}
\usepackage{capt-of}

\usepackage[table]{xcolor}
\usepackage[nocompress]{cite}
\usepackage{cite}

\title{Image Fusion Transformer}

\name{Vibashan VS, Jeya~Maria~Jose~Valanarasu, Poojan Oza, and Vishal M. Patel}
\address{Dept. of Electrical and Computer Engineering, Johns Hopkins University, MD, USA 
\\\(\{\)vvishnu2, jvalana1, poza2, vpatel36\(\}\)@jhu.edu}
\begin{document}
 \ninept
\maketitle
\begin{abstract}
In image fusion, images obtained from different sensors are fused to generate a single image with enhanced information. In recent years,  state-of-the-art methods have adopted Convolution Neural Networks (CNNs) to encode meaningful features for image fusion. Specifically, CNN-based methods perform image fusion by fusing local features. However, they do not consider long-range dependencies that are present in the image. Transformer-based models are designed to overcome this by modelling the long-range dependencies with the help of self-attention mechanism. This motivates us to propose a novel Image Fusion Transformer (IFT) where we develop a transformer-based multi-scale fusion strategy that attends to both local and long-range information (or global context).
The proposed method follows a two-stage training approach.
In the first stage, we train an auto-encoder to extract deep features at multiple scales.
In the second stage, multi-scale features are fused using a Spatio-Transformer (ST) fusion strategy. The ST fusion blocks are comprised of a CNN and a transformer branch which captures local and long-range features, respectively.  Extensive experiments on multiple benchmark datasets show that the proposed method performs better than many competitive fusion algorithms.  Furthermore, we show the effectiveness of the proposed ST fusion strategy with an ablation analysis.
\footnote{The source code is available at:  \href{https://github.com/Vibashan/Image-Fusion-Transformer}{https://github.com/Vibashan/Image-Fusion-Transformer.}}
\end{abstract}
\begin{keywords}
Image fusion, Transformer, CNN, Long-range dependencies, Spatio-Transformer.
\vspace{-4mm}
\end{keywords}
\section{Introduction}
\label{sec:intro}
Image fusion has proved to be critical in many real world applications, e.g., military \cite{ma2019infrared}, computer vision \cite{lopez2018self}, remote sensing \cite{eslami2015developing}, and medical imaging \cite{li2017pixel}.
It refers to combining different images of the same scene to integrate complementary information and generate a single fused image.
For example, the images captured using a visible sensor are rich in fine details like colour, contrast, and texture.
However, visible sensors fail to distinguish between objects and background under poor lighting conditions.
The images obtained using thermal sensors capture salient features that distinguish objects from the background during daytime and nighttime.
However, thermal sensors lack texture and colour space information about the object.
This is because visible sensors work in 300-530 $\mu$m wavelength while thermal sensors work in 8-14 $\mu$m wavelength \cite{tian2015carbon}.
It would be very useful to have a single image that contains complementary information from visible and thermal sensors.
Image fusion specifically tackles this by fusing the complementary attributes from different sources to generate a detailed scene representation \cite{jin2017survey}.

\begin{figure}[t]
\centerline{\includegraphics[width=0.8 \linewidth]{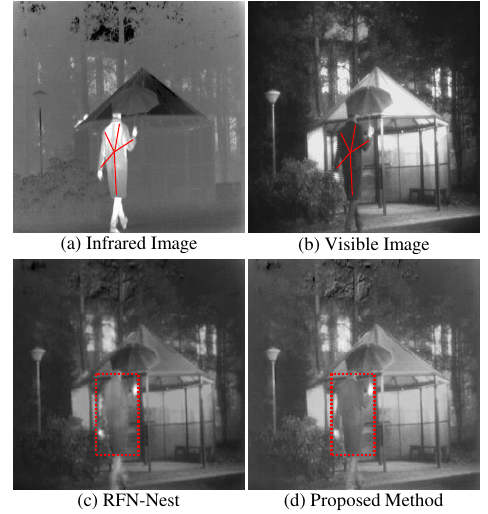}}
\vskip-2.0mm
\caption{(a) Infrared source image, (b) Visible source image. Image fusion results using (c) RFN-Nest \cite{li2021rfn} and proposed method (d) Image Fusion Transformer (IFT). The red line highlights all pixels that belong to the foreground (human). The red box highlights the foreground object fusion by RFN-Nest and IFT. In RFN-Nest, we can observe a gradual decrease of intensity from the top part to the bottom part of the human. This can be explained due to the lack of long-range feature learning capability of CNN-based RFN-Nest. However, IFT does well in encoding these long-range dependencies resulting in the same pixel intensity of the human. Hence, we observe a stark difference in the boundary between foreground and background objects when compared to RFN-Nest.}
\vskip-6.0mm
\label{fig:intro}
\end{figure}

Traditional methods for image fusion include
 sparse representation (SR) based methods  \cite{bin2016efficient,zhang2013dictionary}; multi-scale transformation based methods \cite{hu2017adaptive,he2017infrared}; saliency-based  methods \cite{liu2017infrared} and low-rank representation (LRR) based methods \cite{liu2012robust}.
Even though these methods achieve competitive performance, there exists several shortcomings: 1) They have a poor generalization ability as they rely on handcrafted feature extraction, 2) Dictionary learning in SR and LRR is time-consuming \cite{li2021rfn}, and 3) Different sets of source images require different fusion strategies.

\begin{figure*}[t]
\centerline{\includegraphics[width=0.8 \linewidth]{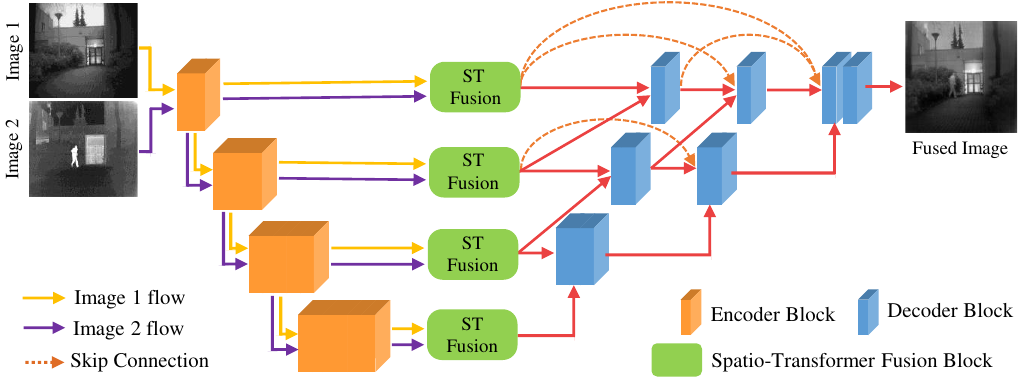}}
\vskip-4.0mm
\caption{Overview of the proposed Image Fusion Transformer (IFT) network. Image 1 and Image 2 are passed through the encoder to obtain multi-scale deep features. These extracted deep features are fused using the Spatio-Transformer (ST) fusion block. Finally, the decoder reconstructs the multi-scale fused features to output a fused image.}
\vskip-5.0mm
\label{fig:archi}
\end{figure*}

Recent image fusion works have explored CNN-based fusion techniques, \cite{li2021rfn}, \cite{xu2020u2fusion}, which outperform traditional ones by overcoming the aforementioned shortcomings.
Though existing CNN-based fusion techniques improve generalization ability by learning local features, they fail to extract long-range dependencies in the images.
This results in the loss of some essential global context that might be useful for an exemplary fused image.
Therefore we argue that integrating local features with long-range dependencies can add global contextual information, which in turn helps to improve fusing performance further.
With this motivation, we propose an Image Fusion Transformer (IFT) with a novel Spatio-Transformer (ST) fusion strategy that effectively learns both local features and long-range information at multiple scales to fuse the complementary information from given images (see Fig.~\ref{fig:intro}).
The main contributions of this work can be summarized as follows:

\begin{itemize}[noitemsep]
    \item  We propose a novel fusion method, called Image Fusion Transformer (IFT), that utilizes both local information and models long-range dependencies to overcome the lack of global contextual understanding that exists in recent image fusion works. 
    \item The proposed method utilizes a novel Spatio-Transformer (ST) fusion strategy, where a spatial CNN branch and a transformer branch are employed to utilize both local and global features to fuse the given images better.
    \item The proposed method is evaluated on multiple fusion benchmark datasets, where we achieve competitive results compared to the existing fusion methods.
\end{itemize}

\section{Related works}
\subsection{Image fusion}

Traditional image fusion methods employed discrete cosine transform (DCT) \cite{cao2014multi}, sparse representation (SR), \cite{bin2016efficient}, principal component analysis (PCA) \cite{kuncheva2013pca}, etc. to extract useful features.
However, these feature extraction methods lack generalizability. Moreover, images captured from different sources require different fusion strategies.
As a result, traditional fusion strategies are designed in a source-specific manner.
Compared to traditional methods, deep learning based methods have shown promising improvement in computer vision tasks such as classification \cite{simonyan2014very,he2016deep,dosovitskiy2020image}, segmentation \cite{ronneberger2015u,vs2022target,valanarasu2022fly} and detection \cite{girshick2015fast,he2017mask,vs2022instance,vs2022towards,vs2022mixture}. Motivated by this, to overcome these issues with the traditional methods for image fusion, the deep learning-based approaches were explored.
 
Li \cite{li2018infrared} proposed a technique where at first, the visible and thermal images are decomposed. Later, they perform fusion using the decomposed images by average feature fusion and deep learning-based feature fusion. 
Another work proposed by Li  \cite{li2018densefuse} uses a fully convolution-based model to fuse the visible and thermal images.
Here, features from source images are extracted using a DenseNet encoder and fused using a CNN-based fusion layer.
A CNN-based decoder is then used to get the fused image.
Li  \cite{li2021rfn} further extended their previous work \cite{li2018densefuse} to an end-to-end fusion strategy for multi-scale deep features minimizing the proposed detail and feature loss.
Xu  \cite{xu2020u2fusion} proposed a unified unsupervised end-to-end framework that tackles the fusion problem by integrating it with continual learning.
However, all of these methods focus on learning spatial local features between source images and do not consider the long-range dependencies present within the source images.
In this work we explore extracting long-range features in addition to the local features to enhance the fusion quality further.

\subsection{Transformers}
Transformer model architecture was first proposed by Vaswani  \cite{vaswani2017attention} and has been proven to be extremely important in Natural Language Processing (NLP) literature over the years. The success of transformer-based models can be attributed to their ability to capture better long-range information compared to recurrent neural networks and CNNs.
Motivated by their success, Dosovitskiy proposed a Vision Transformer (ViT) \cite{dosovitskiy2020image} for image classification.
This has sparked a significant interest in developing transformer-based methods for vision problems like object detection \cite{carion2020end} and  segmentation \cite{valanarasu2021medical}. Hence, in this work, we also exploit a transformer-based architecture to obtain improved image fusion performance by enabling the model to encode long-range dependencies from the images.

\section{Proposed method}
\subsection{Image Fusion Transformer (IFT)}

The proposed Image Fusion Transformer (IFT) is a fusion network that takes in input source images and generates an enhanced fused image.
IFT consists of three parts: encoder network, Spatio-Transformer (ST) fusion network and a nested decoder network as illustrated in Fig. \ref{fig:archi}.
The encoder network consists of four encoder blocks, where each encoder block contains a convolution layer with kernel size $3 \times 3$ followed by ReLU and max-pooling operation.
For a given source input, we extract deep features at multiple-scales from each convolution block of the encoder network.
These extracted features from both images are then fused at multiple scales using the ST fusion network.
The ST fusion network consists of a spatial branch and a transformer branch.
The spatial branch consists of conv layers and a bottleneck layer to capture local features.
The transformer branch consists of an axial attention-based transformer block to capture long-range dependencies (or global context).
Finally, we obtain the fused image by training the nested decoder network with the fused features as an input.
The decoder network is based on the RFN-Nest \cite{li2021rfn} architecture.

\subsection{Self-attention and axial-attention}
Self-attention is an attention mechanism that relates different tokens of a single sequence in order to compute a representation of the same sequence.
Let $x \in \mathbb{R}^{C_{in} \times H \times W}$  and $y \in \mathbb{R}^{C_{out} \times H \times W} $ be the input and output features where $C_{in}$ and $C_{out}$ are the number of input and output channels, respectively and $H$ and $W$ correspond to height and width, respectively.
The output $y$ is computed as follows:
\setlength{\belowdisplayskip}{0pt} \setlength{\belowdisplayshortskip}{0pt}
\setlength{\abovedisplayskip}{0pt} \setlength{\abovedisplayshortskip}{0pt}
\begin{equation}
\label{eq:self}
y_{ij} = \sum_{h=1}^{H}\sum_{w=1}^{W}\text{softmax}(q_{ij}^{T}k_{hw})v_{hw},
\end{equation}
where $q_{ij}$, $k_{ij}$ and $v_{ij}$ are query, key and value at any arbitrary location $i \in \{1,..,H\}$ and  $j \in \{1,..,W\}$ and are computed as $q = W_{Q}x$, $k = W_{K}x$ and $v = W_{V}x$, respectively. 
From Eq. \ref{eq:self}, we can infer that self-attention computes long-range affinities throughout the entire feature map unlike CNN.
However, this self-attention mechanism is computationally expensive due to its quadratic complexity. 

Hence, we employ the axial attention mechanism \cite{ho2019axial} which is computationally more efficient.
Specifically, in axial attention, self-attention is first performed over the feature map height axis and then over the width axis, thus reducing computational complexity.
Moreover, Wang \cite{wang2020axial}, proposed a learnable positional embedding to axial attention query, key and value to make the affinities sensitive to the positional information.
These positional embeddings are parameters that are learnt jointly during training.
Therefore, for a given input $x$, the self-attention along the height axis can be computed as:
\begin{equation}
\label{eq:axial}
y_{ij} = \sum_{h=1}^{H}softmax(q_{ij}^{T}k_{ih} + q_{ij}^{T}r_{ih}^{q} + k_{ij}^{T}r_{ih}^{k})(v_{ih} + v_{ih})
\end{equation}
where $r^{q}$, $r^{k}$, $r^{v}$ $\in \mathbb{R}^{H \times H}$ are the positional embedding for height axis.
For axial attention, we compute Eq. \ref{eq:axial} along both height and width axis, which provides an efficient self-attention model.

\begin{figure}[!h]
\centerline{\includegraphics[width=0.85 \linewidth]{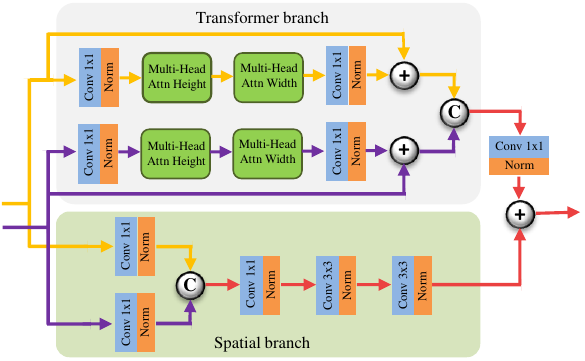}}
\caption{The feed-forward path for Spatio-Transformer (ST) fusion mechanism. The encoded feature maps from Image 1 and Image 2 are fed to the spatial branch and the transformer branch. The spatial branch extracts fine local features while the transformer branch extracts long-range features.}
\vskip-7.0mm
\label{fig:axial}
\end{figure}

\subsection{Spatio-Transformer (ST) fusion strategy}

The proposed ST fusion block consists of two branches: the spatial and transformer branches.
In the spatial branch, we use a conv block and a bottleneck layer to capture local features.
In the transformer branch, we use axial attention to learn global-contextual features by modelling long-range dependencies through the self-attention mechanism.
We add these two features to obtain a fused feature map containing enhanced local and global-context information.
Moreover, we applied our ST fusion strategy at multiple scales and then forwarded it to the decoder network to obtain the final fused image.
ST fusion block is illustrated in Fig. \ref{fig:axial}.

\begin{figure*}[!h]
\centerline{\includegraphics[width=1.0\linewidth]{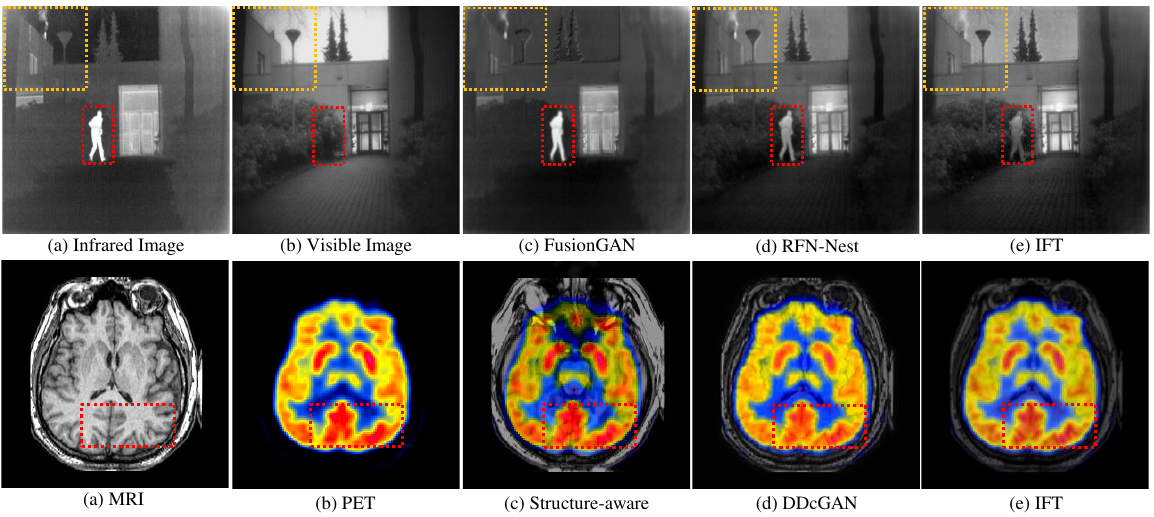}}

\caption{Qualitative comparison: (1) $Top:$ Infrared/Visible; (2) $Bottom:$ MRI/PET. Other fusion methods do not focus on modeling long-range dependencies, while our proposed method encodes both local and global-context, resulting in a better fusion.}
\vskip-3.0mm
\label{fig:quali}
\end{figure*}

\subsection{Loss function}

The proposed method is trained to preserve fine structural details and retain the salient foreground and background details.
The overall training objective to train IFT, denoted as $L_{fuse}$, can be given as:
\begin{eqnarray}\label{equ:loss-rfn}
  	L_{fuse}=L_{feat} + \alpha L_{det}, 
\end{eqnarray}
where $L_{det}$ is the structural similarity loss, which is computed as follows 
\begin{eqnarray}\label{equ:loss-detail}
  	L_{det} = 1 - SSIM(O, I)
\end{eqnarray}
where $O$ and $I$ are the fused and input source image, respectively.
Also, $SSIM(.)$ measures structural similarity.
If $SSIM(O, I)$ tends to 1, then the fused image retains most of the structural details from the source images and vice versa.
The feature similarity loss $L_{feat}$ is calculated as follows 
\begin{eqnarray}\label{equ:loss-detail}
  	L_{feat}=\sum_{m=1}^M w_1||\Phi_f^m - (w_{I1}\Phi_{I1}^m+w_{I2}\Phi_{I2}^m)||_F^2,
\end{eqnarray}
where $M$ is the number scales at which deep features are extracted; $f, I1, I2$ denote fused image, input source 1 image and input source 2 image, respectively.
Also, $w_1, w_{I1}, w_{I2}$ are trade-off parameters to balance the loss magnitude.
$\Phi_f^m$ is the fused feature map while $\Phi_{I1}$ and $\Phi_{I2}$ correspond to the encoded feature maps of the input source 1 and input source 2 images, respectively.
This loss constrains the fused deep features to preserve salient structures, thus enhancing the fused feature space to learn more salient features and preserve fine details.
Here, $\alpha$ is a hyperparameter.

\section{Experiments and results}

\noindent {\bf{Implementation details.}} For visible and infrared fusion, we train our model on 80000 pairs of visible and infrared images in the KAIST dataset.
We test on 21 pairs of visible and infrared images in the TNO Human Factors dataset during testing.
Further, we follow RFN-Nest experimental setup by resizing the images to $256\times 256$ and setting hyperparameters $w_{I1}, w_{I2}, w_1, \alpha$ equal to 6, 3, 100, 700.
For all experiments, we set the learning rate, epoch, and batch size equal to $10^{-4}$, 4 and 2, respectively. 

For the experiment with MRI and PET images, the network is trained on 9981 cropped patches with image pairs obtained from the Harvard MRI and PET datasets. The trained model is evaluated on 20 pairs of MRI and PET images sampled from the Harvard MRI and PET image fusion dataset.
During training,  we resize the images to $84 \times 84$ and convert the PET images to IHS scale to fuse the $I$ channel with an MRI image.
For all experiments, we set the learning rate, epoch, and batch size equal to $10^{-4}$, 4 and 2, respectively.


\begin{table}[h]
\centering
\caption{Quantitative results on 21 pairs of infrared and visible images. En: Entropy, SCD: Sum of the correlations of differences, MI: Mutual Information, MS-SSIM: Multi-scale structural similarity.}
\resizebox{0.95\linewidth}{!}{\begin{tabular}{lcccccc}
\hline
\textbf{Methods}          & \textbf{En} \cite{roberts2008assessment} & \textbf{SCD} \cite{aslantas2015new} & \textbf{MI} \cite{qu2002information} &  \textbf{MS-SSIM} \cite{ma2015perceptual} \\ \hline
DCHWT \cite{kumar2013multifocus}    & 6.5677     & 1.6099     & 13.1355         & 0.8432          \\ \hline
ConvSR \cite{liu2016image}    & 6.2586     & 1.6482     & 12.5173          & 0.9028          \\ \hline
VggML \cite{li2018infrared}    & 6.1826     & 1.6352     & 12.3652          & 0.8747          \\ \hline
DenseFuse \cite{li2018densefuse} & 6.6715     & 1.8350     & 13.3431          & \textbf{0.9289}          \\ \hline
IFCNN  \cite{zhang2020ifcnn}   & 6.5954     & 1.7137     & 13.1909          & 0.9052          \\ \hline
NestFuse \cite{li2020nestfuse} & 6.9197     & 1.7335     & 13.8394         & 0.8624          \\ \hline
FusionGan \cite{ma2019fusiongan} & 6.3628     & 1.4568     & 12.7257          & 0.7318          \\ \hline
U2Fusion \cite{xu2020u2fusion} & 6.7570     & 1.7983    & 13.5141         & 0.9253          \\ \hline
RFN-Nest \cite{li2021rfn} & 6.8413     & \textbf{1.8367}     & 13.6826          & 0.9145          \\ \hline
\textbf{IFT}  & \textbf{6.9862}     & 1.7818     & \textbf{13.9725}         & 0.8606          \\ \hline
\end{tabular}}
\vskip -10 pt
\label{tab:ir}
\end{table}

\noindent {\bf{Infrared and visible image fusion.}} From Table \ref{tab:ir}, we can observe that the proposed method outperforms the existing methods in En and MI metrics. Our method can capture both local and long-range dependencies generating sharper content and preserve most of the visual information compared to other methods. Further, our method produces competitive performance in SCD and MS-SSIM metrics than other methods solely focusing on local image fusion. Qualitative fusion results are illustrated in the top row of Fig. \ref{fig:quali}. The red box highlights the human and the yellow box highlights the reconstruction of fine features. In the red box, we can observe that capturing long-range dependencies results in assigning the same intensity all over the human for IFT compared to other CNN-based methods. In addition, we can observe in the yellow box that our model can reconstruct fine details as it captures both long-range and local information.

\begin{table}[h]
\centering
\vskip -5 pt
\caption{Quantitative results for 20 pairs of MRI and PET image fusion. En: Entropy, SD: Standard Deviation, CC: Correlation Coefficient, MG: Mean Gradient.}
\resizebox{0.95\linewidth}{!}{\begin{tabular}{lcccccc}
\hline
\textbf{Methods}          & \textbf{En} \cite{roberts2008assessment} & \textbf{SD} \cite{rao1997fibre} & \textbf{CC} \cite{ma2020ddcgan} &  \textbf{MG} \cite{ma2020ddcgan}\\ \hline
DCHWT  \cite{kumar2013multifocus}       & 5.7507     & 0.3337     & 0.8510          & 0.0448          \\ \hline
DDCTPCA   \cite{naidu2014hybrid}   & 5.2298     & 0.3351     &  0.9044           & 0.0257         \\ \hline
Structure-Aware  \cite{li2018structure}    & 5.1144     & 0.3426     &  0.8390           & 0.0454          \\ \hline
FusionGAN  \cite{ma2019fusiongan}     & 5.1841     & 0.1741     & 0.8547          & 0.0198          \\ \hline
RCGAN  \cite{li2019coupled}     & 5.6549     & 0.2894    & 0.9000          & 0.0303          \\ \hline
DDcGAN \cite{ma2020ddcgan}    & 5.9787     & \textbf{0.3519}    & 0.9012          & \textbf{0.0471}          \\ \hline
\textbf{IFT}  & \textbf{6.4328}     & 0.3298     & \textbf{0.9463}        & 0.0444          \\ \hline
\end{tabular}}
\label{tab:mri}
\vskip -10 pt
\end{table}

\noindent {\bf{MRI and PET image fusion}} From Table \ref{tab:mri}, we can infer that our method outperforms all the existing methods in Entropy and CC metrics by preserving local and long-range information. In SD and MG metrics, it produces competitive performance compared to the existing techniques. Qualitative fusion results are illustrated in the bottom row of Fig. \ref{fig:quali} and the red box highlights the intensity variation of PET colors in the fused image. From Fig. \ref{fig:quali} we can observe that Structure-aware method \cite{li2018structure} lacks color intensity variations in the fused image; whereas, DDcGAN and IFT exhibit better intensity variations and brighter colors. Moreover, IFT color variation is more similar to PET than DDcGAN, thanks to IFT's ability to encode long-range dependencies. This is inferred from the high performance in Correlation Coefficient (CC) metric from Table \ref{tab:mri}.

\begin{table}[h]
\caption{Ablation study on the ST fusion network for MRI and PET image fusion.}
\huge
\centering
\resizebox{0.95\linewidth}{!}{\begin{tabular}{lcccccc}
\hline
\textbf{Methods}          & \textbf{En} \cite{roberts2008assessment} & \textbf{SD} \cite{rao1997fibre} & \textbf{CC} \cite{ma2020ddcgan} &  \textbf{MG} \cite{ma2020ddcgan}\\ \hline
Spatial      & 5.9813     & 0.2961      & 0.9311         & 0.0382              \\ \hline
Transformer   & 6.1459     & 0.3183     & 0.9392        & 0.0418            \\ \hline
IFT (Spatial+Transformer)      & \textbf{6.4328}     & \textbf{0.3298}     & \textbf{0.9463}        & \textbf{0.0444}          \\ \hline
\end{tabular}}
\label{tab:ablation}
\vskip -10 pt
\end{table}

\noindent {\bf{Ablation study.}} An ablation study is conducted on the ST Fusion branch and the results are reported in Table \ref{tab:ablation}. Spatial-based image fusion performs fusion using only local features, whereas transformer-based image fusion performs fusion operation utilizing long-range dependencies. However, it is crucial to capture both local and long-range features for understanding overall representation, which results in better image fusion.  From Table \ref{tab:ablation}, it is evident that our proposed  ST fusion network outperforms only spatial or transformer-based image fusion in all metrics by capturing both local and long-range dependencies. Hence, this supports our argument that image fusion is improved by integrating long-range dependencies with local features.

\section{Conclusion}
In this work, we proposed an Image Fusion Transformer (IFT) network where we developed a novel Spatio-Transformer (ST) fusion strategy that attends to both local and long-range dependencies.  In the ST fusion strategy, a CNN branch and a transformer branch are introduced to fuse local and global features. The proposed method is evaluated on multiple fusion benchmark datasets where we achieve better results compared to the existing fusion methods. Moreover, we perform an ablation study to show the effectiveness of extracting local and long-range information while doing fusion.


\bibliographystyle{IEEEbib}
{\footnotesize
\bibliography{strings,refs}

\begin{thebibliography}{10}

\bibitem{ma2019infrared}
Jiayi Ma, Yong Ma, and Chang Li,
\newblock ``Infrared and visible image fusion methods and applications: A
  survey,''
\newblock {\em Information Fusion}, vol. 45, pp. 153--178, 2019.

\bibitem{lopez2018self}
Carlos Lopez-Molina, Javier Montero, Humberto Bustince, and Bernard De~Baets,
\newblock ``Self-adapting weighted operators for multiscale gradient fusion,''
\newblock {\em Information Fusion}, vol. 44, pp. 136--146, 2018.

\bibitem{eslami2015developing}
Mehrdad Eslami and Ali Mohammadzadeh,
\newblock ``Developing a spectral-based strategy for urban object detection
  from airborne hyperspectral tir and visible data,''
\newblock {\em IEEE Journal of Selected Topics in Applied Earth Observations
  and Remote Sensing}, vol. 9, no. 5, pp. 1808--1816, 2015.

\bibitem{li2017pixel}
Shutao Li, Xudong Kang, Leyuan Fang, Jianwen Hu, and Haitao Yin,
\newblock ``Pixel-level image fusion: A survey of the state of the art,''
\newblock {\em information Fusion}, vol. 33, pp. 100--112, 2017.

\bibitem{tian2015carbon}
Jian Tian, Yanhua Leng, Zhenhuan Zhao, Yang Xia, Yuanhua Sang, Pin Hao, Jie
  Zhan, Meicheng Li, and Hong Liu,
\newblock ``Carbon quantum dots/hydrogenated tio2 nanobelt heterostructures and
  their broad spectrum photocatalytic properties under uv, visible, and
  near-infrared irradiation,''
\newblock {\em Nano Energy}, vol. 11, pp. 419--427, 2015.

\bibitem{jin2017survey}
Xin Jin, Qian Jiang, Shaowen Yao, Dongming Zhou, Rencan Nie, Jinjin Hai, and
  Kangjian He,
\newblock ``A survey of infrared and visual image fusion methods,''
\newblock {\em Infrared Physics \& Technology}, vol. 85, pp. 478--501, 2017.

\bibitem{li2021rfn}
Hui Li, Xiao-Jun Wu, and Josef Kittler,
\newblock ``Rfn-nest: An end-to-end residual fusion network for infrared and
  visible images,''
\newblock {\em Information Fusion}, vol. 73, pp. 72--86, 2021.

\bibitem{bin2016efficient}
Yang Bin, Yang Chao, and Huang Guoyu,
\newblock ``Efficient image fusion with approximate sparse representation,''
\newblock {\em International Journal of Wavelets, Multiresolution and
  Information Processing}, vol. 14, no. 04, pp. 1650024, 2016.

\bibitem{zhang2013dictionary}
Qiheng Zhang, Yuli Fu, Haifeng Li, and Jian Zou,
\newblock ``Dictionary learning method for joint sparse representation-based
  image fusion,''
\newblock {\em Optical Engineering}, vol. 52, no. 5, pp. 057006, 2013.

\bibitem{hu2017adaptive}
Hai-Miao Hu, Jiawei Wu, Bo~Li, Qiang Guo, and Jin Zheng,
\newblock ``An adaptive fusion algorithm for visible and infrared videos based
  on entropy and the cumulative distribution of gray levels,''
\newblock {\em IEEE Transactions on Multimedia}, vol. 19, no. 12, pp.
  2706--2719, 2017.

\bibitem{he2017infrared}
Kangjian He, Dongming Zhou, Xuejie Zhang, Rencan Nie, Quan Wang, and Xin Jin,
\newblock ``Infrared and visible image fusion based on target extraction in the
  nonsubsampled contourlet transform domain,''
\newblock {\em Journal of Applied Remote Sensing}, vol. 11, no. 1, pp. 015011,
  2017.

\bibitem{liu2017infrared}
CH~Liu, Y~Qi, and WR~Ding,
\newblock ``Infrared and visible image fusion method based on saliency
  detection in sparse domain,''
\newblock {\em Infrared Physics \& Technology}, vol. 83, pp. 94--102, 2017.

\bibitem{liu2012robust}
Guangcan Liu, Zhouchen Lin, Shuicheng Yan, Ju~Sun, Yong Yu, and Yi~Ma,
\newblock ``Robust recovery of subspace structures by low-rank
  representation,''
\newblock {\em IEEE transactions on pattern analysis and machine intelligence},
  vol. 35, no. 1, pp. 171--184, 2012.

\bibitem{xu2020u2fusion}
Han Xu, Jiayi Ma, Junjun Jiang, Xiaojie Guo, and Haibin Ling,
\newblock ``U2fusion: A unified unsupervised image fusion network,''
\newblock {\em IEEE Transactions on Pattern Analysis and Machine Intelligence},
  2020.

\bibitem{cao2014multi}
Liu Cao, Longxu Jin, Hongjiang Tao, Guoning Li, Zhuang Zhuang, and Yanfu Zhang,
\newblock ``Multi-focus image fusion based on spatial frequency in discrete
  cosine transform domain,''
\newblock {\em IEEE signal processing letters}, vol. 22, no. 2, pp. 220--224,
  2014.

\bibitem{kuncheva2013pca}
Ludmila~I Kuncheva and William~J Faithfull,
\newblock ``Pca feature extraction for change detection in multidimensional
  unlabeled data,''
\newblock {\em IEEE transactions on neural networks and learning systems}, vol.
  25, no. 1, pp. 69--80, 2013.

\bibitem{simonyan2014very}
Karen Simonyan and Andrew Zisserman,
\newblock ``Very deep convolutional networks for large-scale image
  recognition,''
\newblock {\em arXiv preprint arXiv:1409.1556}, 2014.

\bibitem{he2016deep}
Kaiming He, Xiangyu Zhang, Shaoqing Ren, and Jian Sun,
\newblock ``Deep residual learning for image recognition,''
\newblock in {\em Proceedings of the IEEE conference on computer vision and
  pattern recognition}, 2016, pp. 770--778.

\bibitem{dosovitskiy2020image}
Alexey Dosovitskiy, Lucas Beyer, Alexander Kolesnikov, Dirk Weissenborn,
  Xiaohua Zhai, Thomas Unterthiner, Mostafa Dehghani, Matthias Minderer, Georg
  Heigold, Sylvain Gelly, et~al.,
\newblock ``An image is worth 16x16 words: Transformers for image recognition
  at scale,''
\newblock {\em arXiv preprint arXiv:2010.11929}, 2020.

\bibitem{ronneberger2015u}
Olaf Ronneberger, Philipp Fischer, and Thomas Brox,
\newblock ``U-net: Convolutional networks for biomedical image segmentation,''
\newblock in {\em International Conference on Medical image computing and
  computer-assisted intervention}. Springer, 2015, pp. 234--241.

\bibitem{vs2022target}
Vibashan VS, Jeya Maria~Jose Valanarasu, and Vishal~M Patel,
\newblock ``Target and task specific source-free domain adaptive image
  segmentation,''
\newblock {\em arXiv preprint arXiv:2203.15792}, 2022.

\bibitem{valanarasu2022fly}
Jeya Maria~Jose Valanarasu, Pengfei Guo, Vibashan VS, and Vishal~M Patel,
\newblock ``On-the-fly test-time adaptation for medical image segmentation,''
\newblock {\em arXiv preprint arXiv:2203.05574}, 2022.

\bibitem{girshick2015fast}
Ross Girshick,
\newblock ``Fast r-cnn,''
\newblock in {\em Proceedings of the IEEE international conference on computer
  vision}, 2015, pp. 1440--1448.

\bibitem{he2017mask}
Kaiming He, Georgia Gkioxari, Piotr Doll{\'a}r, and Ross Girshick,
\newblock ``Mask r-cnn,''
\newblock in {\em Proceedings of the IEEE international conference on computer
  vision}, 2017, pp. 2961--2969.

\bibitem{vs2022instance}
Vibashan VS, Poojan Oza, and Vishal~M Patel,
\newblock ``Instance relation graph guided source-free domain adaptive object
  detection,''
\newblock {\em arXiv preprint arXiv:2203.15793}, 2022.

\bibitem{vs2022towards}
Vibashan VS, Poojan Oza, and Vishal~M Patel,
\newblock ``Towards online domain adaptive object detection,''
\newblock {\em arXiv preprint arXiv:2204.05289}, 2022.

\bibitem{vs2022mixture}
Vibashan Vs, Poojan Oza, Vishwanath~A Sindagi, and Vishal~M Patel,
\newblock ``Mixture of teacher experts for source-free domain adaptive object
  detection,''
\newblock in {\em 2022 IEEE International Conference on Image Processing
  (ICIP)}. IEEE, 2022, pp. 3606--3610.

\bibitem{li2018infrared}
Hui Li, Xiao-Jun Wu, and Josef Kittler,
\newblock ``Infrared and visible image fusion using a deep learning
  framework,''
\newblock in {\em 2018 24th international conference on pattern recognition
  (ICPR)}. IEEE, 2018, pp. 2705--2710.

\bibitem{li2018densefuse}
Hui Li and Xiao-Jun Wu,
\newblock ``Densefuse: A fusion approach to infrared and visible images,''
\newblock {\em IEEE Transactions on Image Processing}, vol. 28, no. 5, pp.
  2614--2623, 2018.

\bibitem{vaswani2017attention}
Ashish Vaswani, Noam Shazeer, Niki Parmar, Jakob Uszkoreit, Llion Jones,
  Aidan~N Gomez, Lukasz Kaiser, and Illia Polosukhin,
\newblock ``Attention is all you need,''
\newblock {\em arXiv preprint arXiv:1706.03762}, 2017.

\bibitem{carion2020end}
Nicolas Carion, Francisco Massa, Gabriel Synnaeve, Nicolas Usunier, Alexander
  Kirillov, and Sergey Zagoruyko,
\newblock ``End-to-end object detection with transformers,''
\newblock in {\em European Conference on Computer Vision}. Springer, 2020, pp.
  213--229.

\bibitem{valanarasu2021medical}
Jeya Maria~Jose Valanarasu, Poojan Oza, Ilker Hacihaliloglu, and Vishal~M
  Patel,
\newblock ``Medical transformer: Gated axial-attention for medical image
  segmentation,''
\newblock {\em arXiv preprint arXiv:2102.10662}, 2021.

\bibitem{ho2019axial}
Jonathan Ho, Nal Kalchbrenner, Dirk Weissenborn, and Tim Salimans,
\newblock ``Axial attention in multidimensional transformers,''
\newblock {\em arXiv preprint arXiv:1912.12180}, 2019.

\bibitem{wang2020axial}
Huiyu Wang, Yukun Zhu, Bradley Green, Hartwig Adam, Alan Yuille, and
  Liang-Chieh Chen,
\newblock ``Axial-deeplab: Stand-alone axial-attention for panoptic
  segmentation,''
\newblock in {\em European Conference on Computer Vision}. Springer, 2020, pp.
  108--126.

\bibitem{roberts2008assessment}
J~Wesley Roberts, Jan~A Van~Aardt, and Fethi~Babikker Ahmed,
\newblock ``Assessment of image fusion procedures using entropy, image quality,
  and multispectral classification,''
\newblock {\em Journal of Applied Remote Sensing}, vol. 2, no. 1, pp. 023522,
  2008.

\bibitem{aslantas2015new}
V~Aslantas and Emre Bendes,
\newblock ``A new image quality metric for image fusion: the sum of the
  correlations of differences,''
\newblock {\em Aeu-international Journal of electronics and communications},
  vol. 69, no. 12, pp. 1890--1896, 2015.

\bibitem{qu2002information}
Guihong Qu, Dali Zhang, and Pingfan Yan,
\newblock ``Information measure for performance of image fusion,''
\newblock {\em Electronics letters}, vol. 38, no. 7, pp. 313--315, 2002.

\bibitem{ma2015perceptual}
Kede Ma, Kai Zeng, and Zhou Wang,
\newblock ``Perceptual quality assessment for multi-exposure image fusion,''
\newblock {\em IEEE Transactions on Image Processing}, vol. 24, no. 11, pp.
  3345--3356, 2015.

\bibitem{kumar2013multifocus}
BK~Shreyamsha Kumar,
\newblock ``Multifocus and multispectral image fusion based on pixel
  significance using discrete cosine harmonic wavelet transform,''
\newblock {\em Signal, Image and Video Processing}, vol. 7, no. 6, pp.
  1125--1143, 2013.

\bibitem{liu2016image}
Yu~Liu, Xun Chen, Rabab~K Ward, and Z~Jane Wang,
\newblock ``Image fusion with convolutional sparse representation,''
\newblock {\em IEEE signal processing letters}, vol. 23, no. 12, pp.
  1882--1886, 2016.

\bibitem{zhang2020ifcnn}
Yu~Zhang, Yu~Liu, Peng Sun, Han Yan, Xiaolin Zhao, and Li~Zhang,
\newblock ``Ifcnn: A general image fusion framework based on convolutional
  neural network,''
\newblock {\em Information Fusion}, vol. 54, pp. 99--118, 2020.

\bibitem{li2020nestfuse}
Hui Li, Xiao-Jun Wu, and Tariq Durrani,
\newblock ``Nestfuse: An infrared and visible image fusion architecture based
  on nest connection and spatial/channel attention models,''
\newblock {\em IEEE Transactions on Instrumentation and Measurement}, vol. 69,
  no. 12, pp. 9645--9656, 2020.

\bibitem{ma2019fusiongan}
Jiayi Ma, Wei Yu, Pengwei Liang, Chang Li, and Junjun Jiang,
\newblock ``Fusiongan: A generative adversarial network for infrared and
  visible image fusion,''
\newblock {\em Information Fusion}, vol. 48, pp. 11--26, 2019.

\bibitem{rao1997fibre}
Yun-Jiang Rao,
\newblock ``In-fibre bragg grating sensors,''
\newblock {\em Measurement science and technology}, vol. 8, no. 4, pp. 355,
  1997.

\bibitem{ma2020ddcgan}
Jiayi Ma, Han Xu, Junjun Jiang, Xiaoguang Mei, and Xiao-Ping Zhang,
\newblock ``Ddcgan: A dual-discriminator conditional generative adversarial
  network for multi-resolution image fusion,''
\newblock {\em IEEE Transactions on Image Processing}, vol. 29, pp. 4980--4995,
  2020.

\bibitem{naidu2014hybrid}
VPS Naidu,
\newblock ``Hybrid ddct-pca based multi sensor image fusion,''
\newblock {\em Journal of Optics}, vol. 43, no. 1, pp. 48--61, 2014.

\bibitem{li2018structure}
Wen Li, Yuange Xie, Haole Zhou, Ying Han, and Kun Zhan,
\newblock ``Structure-aware image fusion,''
\newblock {\em Optik}, vol. 172, pp. 1--11, 2018.

\bibitem{li2019coupled}
Qilei Li, Lu~Lu, Zhen Li, Wei Wu, Zheng Liu, Gwanggil Jeon, and Xiaomin Yang,
\newblock ``Coupled gan with relativistic discriminators for infrared and
  visible images fusion,''
\newblock {\em IEEE Sensors Journal}, 2019.

\end{thebibliography}
}

\end{document}